# CHANGE DETECTION USING SYNTHETIC APERTURE RADAR VIDEOS

Hasara Maithree, Dilan Dinushka and Adeesha Wijayasiri

Department of Computer Science and Engineering,
University of Moratuwa, Moratuwa, Sri Lanka

*ABSTRACT*

*Many researches have been carried out for change detection using temporal SAR images. In this paper an algorithm for change detection using SAR videos has been proposed. There are various challenges related to SAR videos such as high level of speckle noise, rotation of SAR image frames of the video around a particular axis due to the circular movement of airborne vehicle, non-uniform back scattering of SAR pulses. Hence conventional change detection algorithms used for optical videos and SAR temporal images cannot be directly utilized for SAR videos. We propose an algorithm which is a combination of optical flow calculation using Lucas Kanade (LK) method and blob detection. The developed method follows a four steps approach: image filtering and enhancement, applying LK method, blob analysis and combining LK method with blob analysis. The performance of the developed approach was tested on SAR videos available on Sandia National Laboratories website and SAR videos generated by a SAR simulator.*

## 1. INTRODUCTION

Synthetic Aperture Radar (SAR) is an important modality for remote sensing applications since it has the capability of generating high resolution images significantly invariant to the climate changes, weather and lighting conditions. Tensor product-based transformation of radar return pulse histories are applied to obtain a spatial representation of target objects. SAR imagery uses the motion of radar antenna over a target region to provide a finer spatial resolution than a normal beam scanning radar [1], [2].

ViSAR is a SAR imaging mode which is utilized to generate images at a higher rate than a conventional SAR, hence can be viewed as a video derived from consequent set of image frames. Since images generated from ViSAR systems have a higher resolution despite the adverse climate changes, these systems can be utilized in many real-world day/nights surveillance and tracking applications [3].

Most of the research that has been carried out, were focused on change detection in temporal images using SAR imagery which means, detecting changes between images that have acquired on different dates. In this paper, we are going to discuss how change detection can be applied on real time Synthetic Aperture Radar (SAR) videos which are generated by aligning consecutive image frames.

We propose a new method for change detection using the combination of Lucas Kanade method and blob detection followed with various pre-processing steps for filtering and image enhancement. Since SAR video generation can be paralleled and can be extended to do in real time [4], applying change detection on SAR videos can be useful for real time surveillance operations in military situations, ship detection, rescue operations in the aftermath of natural disasters, traffic monitoring and searching and tracking for various other applications.

## 2. BACKGROUND AND RELATED WORK

### 2.1. SAR Video Generation

SAR pulse emitter and receiver are located on an airborne platform which travels along a circular path; therefore, it has the effect of covering the same geographical area with different angles which helps to build a complete image of the scene. To derive images from SAR pulse data Frequency domain approaches such as range Doppler imaging and time domain processing algorithms such as Back-propagation were utilized. Due to the support for higher resolution and lesser assumptions about the image, Back-propagation produces better quality constructions compared to frequency domain algorithms [1].

SAR videos are generated by rendering sequence of consecutive SAR pulse reconstructed images. In this paper, a SAR video provided by Sandia Laboratories website and videos generated by our SAR simulator are used for evaluation purposes. Our SAR simulator was developed based on RaySAR [5] to produce circular SAR videos.

### 2.2. Speckle Noise

As a result of the coherent imaging mechanism, SAR images are accompanied by speckle noise unlike optical images. Speckle noise in SAR images are generated as a result of random interference of many elementary reflectors within one resolution cell [6], [7] and is multiplicative in nature. It is observed that speckle noise can affect the quality of the image, image segmentation, classification, extraction of regions of interest and target detection. Hence pre-processing techniques should be applied to reduce the effect of speckle noise. Ideal speckle filter should be adaptable and preserve image statistics, structure of the image and should have simplicity and effectiveness in speckle noise reduction [8].

Various despeckling methods are suggested by researchers and each has its own pros and cons. Converting the nature of speckle noise from multiplicative to additive can be done via log transformation. However, the drawback of log transformation is that it changes the statistical characteristics of the speckled image. Although it can be recovered using inverse log transformation operation, still there remains issues [9].

Mean filter, Median filter, Lee filter, Refined Lee filter and Lee Sigma filter are the most common and simple despeckling techniques used in SAR imagery [8], [10]. While mean filter smooths out the image, it also smooths the edges of the image [10]. As mean filter does not consider the homogeneous, flat areas of the image, it shows a low-edge preservation [11]. Boxcar filter which is a type of mean filter has been used for Polarimetric SAR classification of agricultural region [12]. Segregated noise points in the image can be despeckled using Median filter [13]. Lee filter which uses the Minimum Mean Square Error filter principle reduces the speckle to a considerable level, however edges of the image also get blurred. As an improvement to this Lee filter, Refined Lee filter approach can preserve the edges of the image while reducing the noise [10]

## 2.3. Image Registration

Image registration can be defined as the process of transforming different sets of data into one coordinate system, also can be interpreted as the process of aligning two or more images having a geometrical overlapping area. Images can be taken at different times, from different angles or from different sensors [14]. The procedure for registering two remote sensing images have several steps: (a) Pre-processing, (b) Feature Selection, (c) Feature Correspondence, (d) Determination of a transformation function and (e) Resampling. [15]. Point matching problem of image registration was addressed by implementing a genetic algorithm approach which employed a nearest neighbourhood-based method [16]. To rectify and correct the rotation and translation of SAR images, edge feature consensus method was incorporated for coarse to fine registration [17]. Straight lines, junctions and T-points are significantly visible in man-made structures [14]. For registration of city images, straight lines are considered as an important feature. A hybrid approach of combining area-based techniques with feature-based techniques was often employed as an effective solution for satellite image registration [18], [19].

Image registration process is based on identifying the control points which precisely locates the corresponding image coordinates of the images need to be aligned. Control points can be selected manually or by semi or fully automatic techniques. However manual selection of control points is time consuming and not applicable for near real time image registration. There have been discussed various methods for semi or fully automatic control points selection. Existing automated methods fall into two categories which are feature based or area-based approach. In feature-based methods, features such as curvatures, moments, areas, contour lines or line segments are used to perform registration. Since these features are in variant of climate changes and grey scale changes, feature based methods have shown comparatively accurate results in registration. However, these methods are effective only when features are well presented and preserved. Therefore, area-based image registration methods are still widely used in registration [20].

## 2.4. Change Detection

Change detection techniques can be discussed under both optical and radar imagery. Further these techniques can be categorized under pixel and object-based techniques [21]. By grouping neighbouring pixels based on spectral, textural and edge features, Object Based Change Detection techniques utilize the rich features-based format for analysing pixel regions [22]. Object based techniques are performed by segmenting the image into homogeneous sections based on the spectral aspects of the image. Pixel based approach is done through a pixel-by-pixel comparison. Further this technique can be divided into supervised and unsupervised approach. In supervised change detection, multi temporal images are classified based on external information which is known as the post-classification approach [23]. Volpi et al. (2013) discusses about a supervised approach which is not based on post classification method. They have done the multi temporal image classification using combination of support vector machine (SVM) and spectral properties [24]. The main issue of the supervised change is the need of external information about the imagery [23].

Compared to Supervised Change Detection techniques, Unsupervised change detection techniques use information only included in the imagery itself. Using this technique, image frame can be identified either as changed or unchanged. Thus, unsupervised techniques only include two classes [25]. These techniques can be explained in multiple steps. As the initial step, image pre-processing is performed to reduce the speckle noise. Then a difference image is created from image pixel-by-pixel subtraction or any other method which detects the different pixel values which are deviated from the defined threshold value. The difference image is used to create the change detection map. And the map demonstrates the changed and unchanged areas comparing each neighbouring frame [26], [27]. Threshold on image or a histogram can be applied in this method. The main drawback of Unsupervised Change Detection, as explained by Yousif et al. (2013) is that, it does not elaborate and specify about the change that has been taken place. However, pixel-based methods are sensitive to "salt and pepper" (black and white intermingling of images) noise [23].

Another aspect that should be taken into consideration is the dynamic background of the image sequence. Since the video frame has a dynamic circular moving background, first and foremost background modelling techniques should be done. Background modelling and subtraction which has been widely used for change detection and target detection is prone to false alarms in dynamic background since the background model contains only temporal features. Temporal only methods lack the knowledge of the neighbourhood pixels of the concerned pixel. Thus, it will conclude dynamic background also as a moving object which will cause a false alarm. A new pixel wise nonparametric change detection algorithm has been proposed. The background is modelled by spatiotemporal model using sequences of frames and sampling them in neighbourhood region randomly. Thus, this model contains both spatial and temporal knowledge about the background which leads to better performance in change detection in dynamic background [28].

Even though majority of the papers have discussed identifying temporal changes of SAR imagery which are acquired in different dates, the problem that we address, requires identifying changes in near real time manner and, instead of images, we are dealing with SAR videos which are generated from sequence of images. In order to tackle this problem, key challenges that we have identified, are as follows. 1. Rotation of video frames 2. Speckle noise and other noises which lead to false positives 3. Near real time change detection

## 3. METHOD

Proposed solution for detecting changes is divided into several steps and presented in this section. As a pre-processing step, we use image filtering and enhancement techniques to sharpen the desired objects and to distinguish them from the dynamic background. As the change detection methodology, we use a combination of Lucas Kanade method (LK method) and blob detection for identifying the interesting changes.

### 3.1. Image Filtering/ Noise Reduction

1) Unsharp Masking was used to sharpen the image frames. Consecutive video frames in Figure 1 demonstrate the blurriness of the interesting changes which are the vehicle movements (denoted by a red circle). They are visible as shadow movements to naked eye. Hence the requirement is to sharpen the frames to highlight the moving objects (vehicles) before binarization. Technique of Unsharp masking which is performed as a difference of Gaussian operations can be explained in the following steps.
   a) Blurred image is the exact opposite of sharpened image. By duplicating the original frame and performing Gaussian blurring, blurred frame can be obtained.
   b) Subtracting the blurred image obtained in step a) from the original image to obtain the image with enhanced edges and sharpness (unsharp mask).
   c) Duplicating original image and increasing contrast to obtain high contrast version of the original image.
   d) For each pixel of the unsharp mask, if the luminosity is 100%-pixel value of high contrast version image is used, if 0%, value from the original image is used. Otherwise if the luminosity is in between 0% and 100%, weighted average value of both pixel values of high contrast version image and the original image is used.

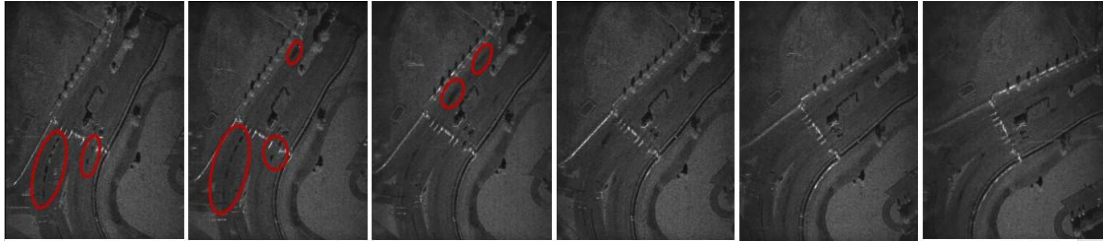

Figure 1. Image frame sequence of a SAR video

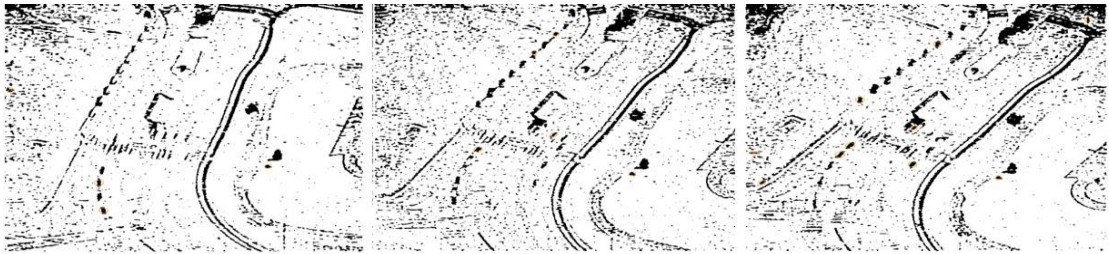

Figure 2. Binarized and unsharp masking added frames of the video.

2) Otsu Binary Conversion

A binary image can be considered as a logical array of 1 s and 0 s. Each pixel of the image can be either black or white. In binary conversion, the intensity range is converted to a 2 level (binary) thresholding. If the pixel value is greater than the threshold, it is replaced by 1, otherwise 0. The Otsu binarization returns a single intensity threshold that separate pixels into two classes, foreground and background. Converting to binary is often advantageous in finding the region of interest for further processing. Moving cars on the road are visible in the enhanced video frames shown in Figure 2.

### 3.2. Rotation Correction

Circular SAR videos rotate due to the method of generation and therefore all the points including interesting and non-interesting points change with each frame. Normal change detection algorithms are incapable in such conditions and therefore an approach to reduce the video rotation was considered. As the first part of the rotation correction, feature point identification was done. For this task several feature extraction methods were tested including Scale Invariant Feature Transform (SIFT) [29], Speeded Up Robust Features (SURF) [30] etc. Algorithms like Harris Corner Detector, Shi-Tomasi Corner Detector was not considered as they do not include any feature descriptors. Oriented Fast Rotated Brief (ORB) algorithm was used as the feature detection and matching algorithm as the comparisons in literature indicated that it is the fastest for mentioned task [31].

After identifying feature points in two consecutive frames in a SAR video, matching feature point was done using Hamming Distance as shown in Figure 3. Then the matched points were sorted according to the distance and filter out the best points.

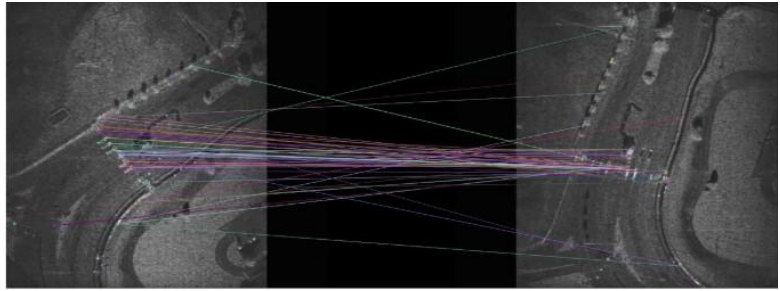

Figure 3. Feature Matching

## 3.3. Change Detection

### 3.3.1. Phase 1 – LK Method

In order to detect interesting changes in a frame such as movements, Lucas Kanade Method for optical flow estimation was used in our application. Optical flow is the motion of objects between consecutive frame as a result of relative movement between the object and the camera. This uses the concept of flow vectors to determine the motion between two subsequent frames of the video. Vectors have both magnitude and direction; hence the flow vector gives an approximation on the amount of deviation of the pixel from current position to the next frame position. We used Sparse optical flow for this purpose which selects sparse set of pixels (interesting features) to calculate its velocity vectors.

LK method is usually used in sparse feature set and our main focus is to find such points from a frame. Usage of dense optical flow calculation methods are not considered as they cannot be used in real time applications. Therefore, as a solution, for each video frame, we choose a uniform distribution of set of fixed points in the video frame and use those points regularly in every optical flow measuring cycle to identify interesting movements. This significantly reduces the computational complexity of the overall optical flow calculation. We can safely assume that interesting movements will pass through one of those selected points in point distribution as the distribution is uniformly spread across the frame. We used a two-threshold mechanism based on motion variance and the angle of deviation to select interesting points from the distribution. If a pixel point has a relative motion or an angle of deviation compared to the pixel points of the neighbourhood, it can be considered as an interesting change. Relative motion vector of the pixel point is obtained by LK method. All objects in the frame are subjected to the rotation of the airborne vehicle and therefore objects which are static have the same angle of rotation, whereas moving objects have a deviation in angle of rotation with respect to the static objects. This angle can be measured using the output of LK method.

Let $(x_1, y_1)$ be a pixel point of the pixel distribution of current frame. Predicted location $(x_2, y_2)$ of the pixel point in the next frame can be derived by LK method. The angle of deviation can be derived as a tangent value ($\theta$) as shown in Figure 4. (a).

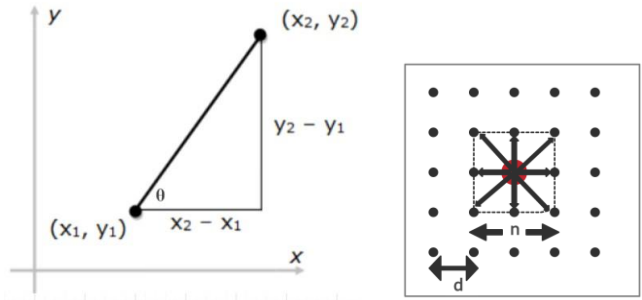

(a) Graphical representation of apparent movement of a pixel. (b) Point distribution of $t^{th}$ frame.

Figure 4. Graphical representation of a pixel and the point distribution.

$$\text{apparent movement} = \sqrt{(x_1-x_2)^2 + (y_1-y_2)^2} \quad (1)$$

$$\tan(\theta) = \frac{y_2 - y_1}{x_2 - x_1} \quad (2)$$

LK method with two threshold methodology to find interesting movements in video frames is explained as follows.

Define two arrays: interesting points array and counter array. Let t be the current frame.

1) Define the point distribution of the $t^{th}$ frame.
   Define a set of points distributed throughout frame. This set of points should be fixed for the frame and will be checked for optical flow changes. As shown in Figure 4. (b), black colour pixels represent the selected pixel distribution of the frame. Pixels are selected such that each pixel is $d$ distance away from another. Assume there are N number of pixel points in the distribution of $t^{th}$ frame.
2) Calculate the optical flow for all N number of pixel points in the distribution.
3) Select interesting points from the point distribution as explained follows: Let $(x_k, y_k)$ be a pixel of the pixel distribution of $t^{th}$ frame. This pixel is represented by the red pixel marked in the Figure 4. (b). Define a window size to derive the neighbourhood of the selected pixel. Let $n \times n$ be the neighbourhood window size. Use LK method to calculate the motion vector of the pixels and to derive the angle of deviation. Let $k_1$, $k_2$ be the thresholds.

   For $k = 1$ to $k = N$ repeat the following:
   For the neighbourhood of $(x_k, y_k)$
   a) Apply LK method to all the pixels in the neighbourhood and derive the average motion of the neighbourhood ($\bar{v}_k$)
   b) Derive the average motion angle of the neighbourhood ($\bar{\theta}_k$).
   c) Derive motion ($v_k$) and the angle ($\theta_k$) of the pixel ($x_k, y_k$).
   d) If $\left|\frac{\bar{v}_k - v_k}{\bar{v}_k}\right| > k_1$ and $\left|\frac{\bar{\theta}_k - \theta_k}{\bar{\theta}_k}\right| > k_2$, mark the point $(x_k, y_k)$ as an interesting point.
   e) If $(x_k, y_k)$ is found interesting,
      i) add $(x_k, y_k)$ to the interesting points array.
      ii) add an entry to the counter array corresponding to the pixel point $(x_k, y_k)$ and initialize its value to zero.

Counter array is to determine whether a selected interesting point in a frame is interesting to all the consecutive frames. The threshold approach of counter array is explained in step 5.

4) If $t \neq 1$,

Let $\bar{V}_t$ be the average sum of the motion vector of all the neighbourhoods of each pixel point in the distribution of frame t,

$$\bar{V}_t = \frac{1}{N}\sum_{k=1}^{N}\bar{v}_k \quad (3)$$

a) Follow steps 1,2 and 3 to find interesting points in the $t^{th}$ frame.
b) Filter already selected points in the interesting points array. Let $\bar{V}_T$ be the average sum of motion vectors from $1^{st}$ to $t = T^{th}$ frame.

$$\bar{V}_T = \frac{1}{T}\sum_{t=1}^{T}\bar{V}_t \quad (4)$$

Let $p_i$ be a point in interesting points array and its motion vector be $v_{pi}$. For each point in interesting point array,
  i) Check the condition:
    A) If $\left|\frac{\bar{V}_T - v_{p_i}}{\bar{V}_T}\right| > k_1$ point is again considered interesting. Initialize its counter back to zero.
    B) Otherwise increase counter value by 1.
  ii) If an interesting point has a counter value larger than the defined threshold value $k_3$, remove it from the interesting points array.
5) Increment to next frame and repeat from step 1.

Storing interesting points of previous frames helps to identify potential actual moving objects. Also, marking them with a counter to determine whether that point is an interesting point for each frame will be used to reduce calculations. The above-mentioned procedure requires three threshold values to be given before the execution: threshold of angle and magnitude of motion vector to determine interesting points, threshold to eliminate false positive interesting points from interesting points array (counter value threshold).

### 3.3.2. Phase 2 – Blob Detection

Blob refers to the group of connected pixels in a binary image and the goal of blob detection is to identify and mark the connected pixel sets in each image.
Blob analysis was used for image segmentation as it can identify pixel clusters with special features which can be interesting changes or object movements.
1) Frame Acquisition: As the SAR video is created by sequence of image frame, 1st frame is acquired for applying image enhancement and blob detection.
2) As SAR videos are inherently black and white, RGB colour components are not considered throughout the algorithm.
3) Image enhancement using Histogram Equalization: Histogram Equalization was utilized to adjust image intensities to enhance the contrast.
4) Blob analysis: Blob analysis, which is a fundamental concept in computer vision, can be performed to distinguish pixel clusters in the image from the background. Blob analysis is applied to find the exact objects in the processed video frames. The objects in the frame which have a clear pixel cluster compared to background and noise, can be either static or dynamic. Since the requirement is to identify changes which can be interpreted as dynamic object movements, the separation of static and dynamic blobs is required. This can be achieved by combining LK method with blob detection. If a pixel cluster can be identified as a clear blob based on the constraints defined on the features of the blob and if it has an apparent motion compared to the previous consecutive frames which can be calculated by LK method, the probability of being an interesting change will be increased.

5) Defining thresholds for features of the blob: In order to identify interesting pixel clusters, constraints for blob features were defined as follows.

   a) By area: Area of the blob is the number of pixels included in the blob. This feature was used to remove blobs which were too small. By setting a minimum and maximum range for area of the blobs, possible objects can be traced down. min area - Minimum area should be defined considering the ratio of the size of the object to the size of the geographical area captured by the frame. max area - area of the circle of which radius is defined as the half of the distance between two-pixel points of theselected pixel distribution of the frame (radius = $d/2$, where $d$ is denoted in Figure 4. (b)).
   When generating SAR image frames, objects are modelled by radar pulse reflections from actual objects, hence pixels of the objects usually have higher intensity compared to the background. Since we are interested in dynamic changes (of objects), pixels are filtered by a pixel intensity range between 0 to 125.

   b) By circularity: Circularity of the blob defines how circular the blob is. This circularity can be measured by Heywood Circularity Factor. Since the SAR video frames are high resolution image frames which cover a large geographical area, the ratio between object area and frame size is very small. Hence these objects can be segmented to circular shapes by a threshold of circular factor.

**3.3.3. Phase 3 – Combining LK method with Blob Detection**

In order to determine whether an actual change is occurred, the combination of LK method and blob detection is applied in an iterative manner for each frame of the video. The methodology is explained as follows.

1) Interesting points array and the detected blobs of the current frame are considered. Let t be the current frame number.
   a) Let ($x_t$, $y_t$) be an interesting pixel point of $t^{th}$ frame and assume there are $N$ number of blobs detected in $t^{th}$ frame.
   b) Let point ($x_n$, $y_n$) be the middle point of an arbitrary blob as denoted by the red pixel in Figure 5. Let $r_n$ be the radius of the detected blob. Consider the area of the square which is drawn $r_n$ distance away from horizontal and vertical directions from ($x_n$, $y_n$), as shown in Figure 5. Let $a_n$ be the array of pixel points which are within the drawn square for ($x_n$, $y_n$)
   c) For $n = 1$ to $n = N$ repeat the following
      i) Derive ($x_n$, $y_n$), $r_n$, $a_n$
      ii) For all points in $a_n$ check whether ($x_t$, $y_t$) is found. If found, mark that point as an interesting change which can be tracked. End the loop.
      iii) Otherwise, increment n by 1, to consider the next blob and repeat the loop.
   d) Repeat steps (a), (b) and (c) for all the interesting points of the current frame.

2) Repeat step 1 for all the frames of the video.

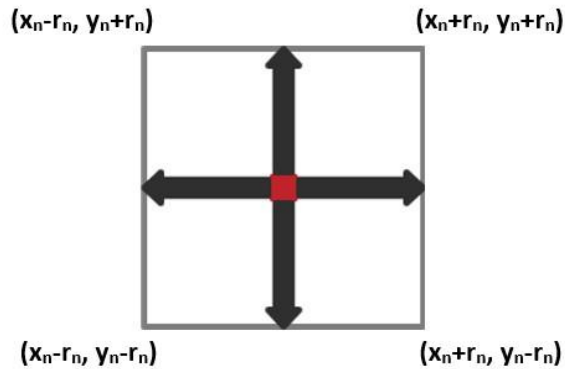

Figure 5. Square of area 4. $(r_n)^2$, drawn using $(x_n, y_n)$ as the centre

## 4. EXPERIMENTS AND RESULTS

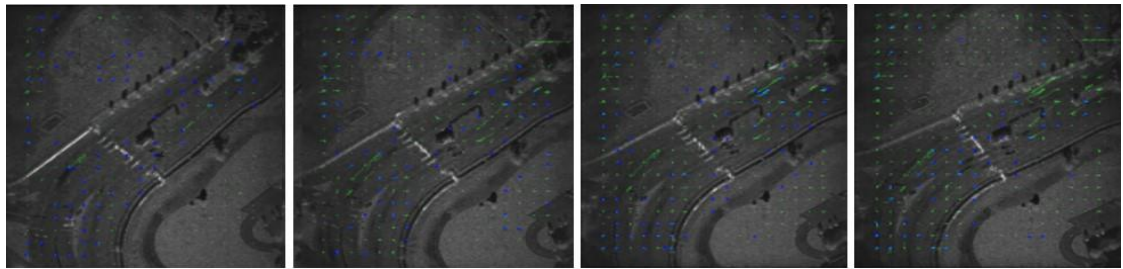

Figure 6. Calculated interesting points and optical flow are shown on the image frames of the SAR video.

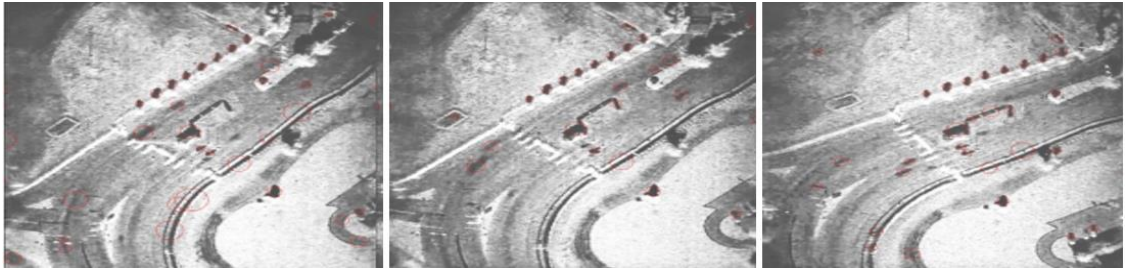

Figure 7. Blobs identified in consecutive image frames of the SAR video.

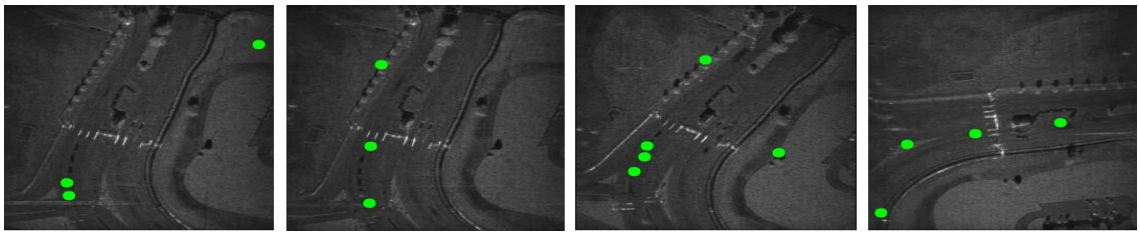

Figure 8. Detected changes

Implementation of our change detection algorithm was tested on SAR videos which were publicly available on Sandia Laboratories Website and videos that were generated by our SAR simulator. Figure 6 shows the applying of LK method for the video taken from Sandia laboratories website. The apparent motion of the pixels which exceeded the two thresholds were marked in blue colour, which we defined as interesting points. Apparent motion of interesting points was tracked by LK method throughout the video sequence and marked in green colour lines as shown in Figure 6 and it is visible that, movements of the vehicles were traced. Majority of the lines drawn were within the road. Hence it can be assumed that interesting movements were captured as moving objects in the frame are vehicles which were driven on the road. Also it is visible that majority of the interesting points were marked around the road (blue points in Figure 6), which means algorithm was able to capture actual interesting points.

Figure 7 shows how blobs are detected in the prepossessed frame using histogram equalization. Blobs are marked by red circles. Even though static objects were also detected by blobs, those blobs could be neglected by combining with LK method. The final changes that are detected by the system using both LK method and blob detection is shown in Figure 8.

Figure 9 represents image frame sequences of SAR videos generated by our SAR Simulator. The simulated video is consisted of moving vehicles on the roads as dynamic objects and buildings and trees as static objects. As shown in Figure 10 moving vehicles were detected by the algorithm.

Even though actual interesting changes were detected in the videos, static objects were captured as interesting changes in some frames, hence caused few false positives. Parameters of blob detection, window size for optical flow calculation of the neighbourhood and pixel step count for determining the pixel distribution of the frame should be further optimized to achieve a higher accuracy.

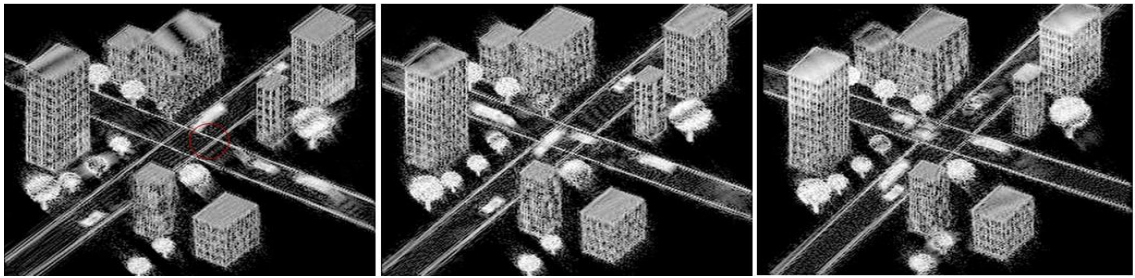

Figure 9. Consecutive image frames of a SAR video generated by SAR Simulator.

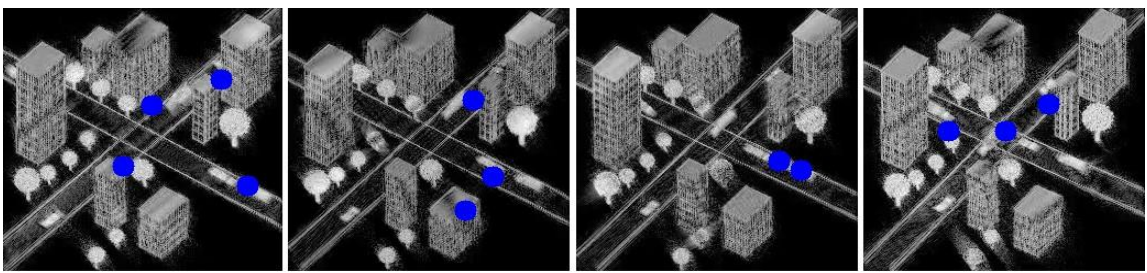

Figure 10. Interesting changes marked in blue points.

## 5. CONCLUSION AND FUTURE WORK

In this paper, an algorithm based on optical flow and blob detection was introduced for change detection in SAR videos. Compared to temporal SAR images, there are unique problems that should be addressed in SAR videos. SAR video frames are subjected to rotation around a particular axis and speckle noise is inherent. Hence when determining an algorithm for change detection of SAR videos, the typical change detection algorithms which were used for optical videos and temporal SAR images could not be directly utilized. The approach proposed in this paper is a modification done by combining LK method and blob detection.

In future work, we hope to do further improvements to increase the accuracy of detection of changes such as defining new methodologies to obtain better threshold values/parameters. These values can be used to keep track of identified interesting changes using a tracking algorithm throughout the video. Also, the algorithms can be implemented parallel to reduce the time complexity. Since SAR videos can be generated using GPUs in real time, reducing the time complexity will be useful for detecting changes in real time.

### ACKNOWLEDGEMENTS

This research was supported by Computer Science Engineering Department of University of Moratuwa. Therefore, we would like to acknowledge all the anonymous colleagues and academic members who gave their valuable insights for this research.

**AUTHORS**

**Hasara Maithree**
Final year undergraduate of Department of Computer Science and Engineering, University of Moratuwa

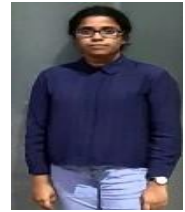

**Dilan Dinushka**
Final year undergraduate of Department of Computer Science and Engineering, University of Moratuwa

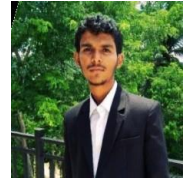

**Dr. Adeesha Wijayasiri**
Lecturer, Department of Computer Science and Engineering, University of Moratuwa

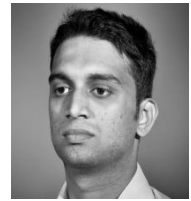